\documentclass[letterpaper, 10 pt, conference]{ieeeconf}  % Comment this line out if you need a4paper

\IEEEoverridecommandlockouts                              % This command is only needed if 
\usepackage{amsmath} % assumes amsmath package installed
\usepackage{amssymb}  % assumes amsmath package installed
\usepackage{subfigure, graphicx, booktabs, enumerate, url, pifont, textcomp, cite}
\usepackage{nth}
\usepackage{bm}

\title{\LARGE \bf
Toward Autonomous Rotation-Aware Unmanned Aerial Grasping
}

\author{Shijie Lin$^{1*}$, Jinwang Wang$^{1*}$, Wen Yang$^{1}$ and Guisong Xia$^{2}$% <-this % stops a space
\thanks{The work was supported in part by the Funds for Creative Research Groups of Natural Science Foundation of Hubei Province under Grant 2018CFA006.}% <-this % stops a space
\thanks{*The first two authors contribute equally to this letter.}
\thanks{$^{1}$Shijie Lin, Jinwang Wang and Wen Yang are with the School of Electronic Information, Wuhan University Wuhan 430072, China {\tt\small linshijie@whu.edu.cn; jwwangchn@whu.edu.cn; yangwen@whu.edu.cn.}}%
\thanks{$^{2}$Guisong Xia are with the State Key Laboratory of Information
Engineering, Surveying, Mapping and Remote Sensing (LIESMARS),
Wuhan University, Wuhan, 430079 China {\tt\small guisong.xia@whu.edu.cn.}}
}

\begin{document}

\maketitle
\thispagestyle{empty}
\pagestyle{empty}

\begin{abstract}
    \label{section:abstract}
    Autonomous Unmanned Aerial Manipulators (UAMs) have shown promising potentials to transform passive sensing missions into active 3-dimension interactive missions, but they still suffer from some difficulties impeding their wide applications, such as target detection and stabilization.
    This letter presents a vision-based autonomous UAM with a 3DoF robotic arm for rotational grasping, with a compensation on displacement for center of gravity.
    First, the hardware, software architecture and state estimation methods are detailed.
    All the mechanical designs are fully provided as open-source hardware for the reuse by the community.
    Then, we analyze the flow distribution generated by rotors and plan the robotic arm's motion based on this analysis.
    Next, a novel detection approach called Rotation-SqueezeDet is proposed to enable rotation-aware grasping, which can give the target position and rotation angle in near real-time on Jetson TX2.
    Finally, the effectiveness of the proposed scheme is validated in multiple experimental trials, highlighting it's applicability of autonomous aerial grasping in GPS-denied environments.

\end{abstract}

\section{INTRODUCTION}
\label{section:introduction}
\begin{figure}
	\centering
	\subfigure[ ]
	{
		\label{fig:grasping}
		\includegraphics[width=0.93\linewidth]{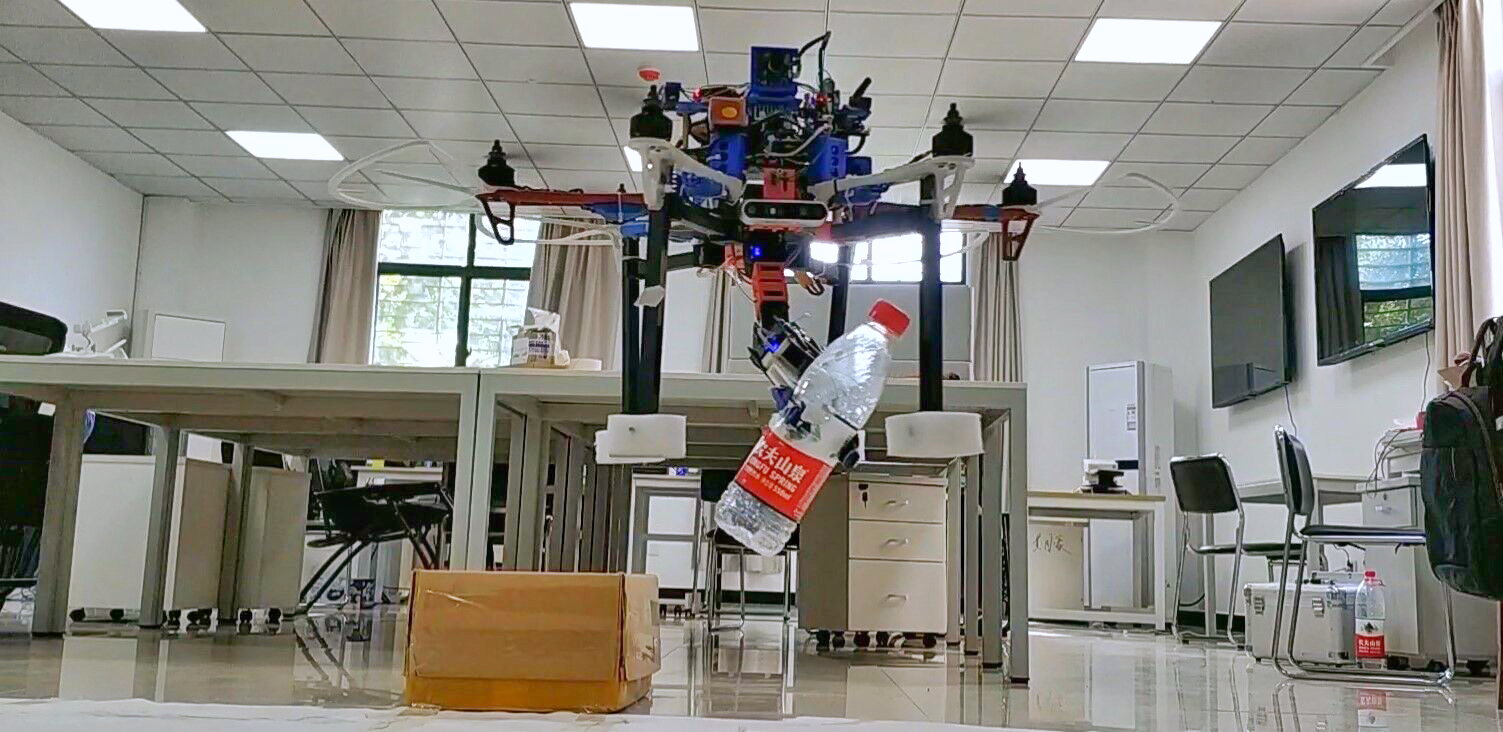}
	}
	\subfigure[ ]
	{
		\label{fig:UAV-BD-Samples}
		\includegraphics[width=0.28\linewidth]{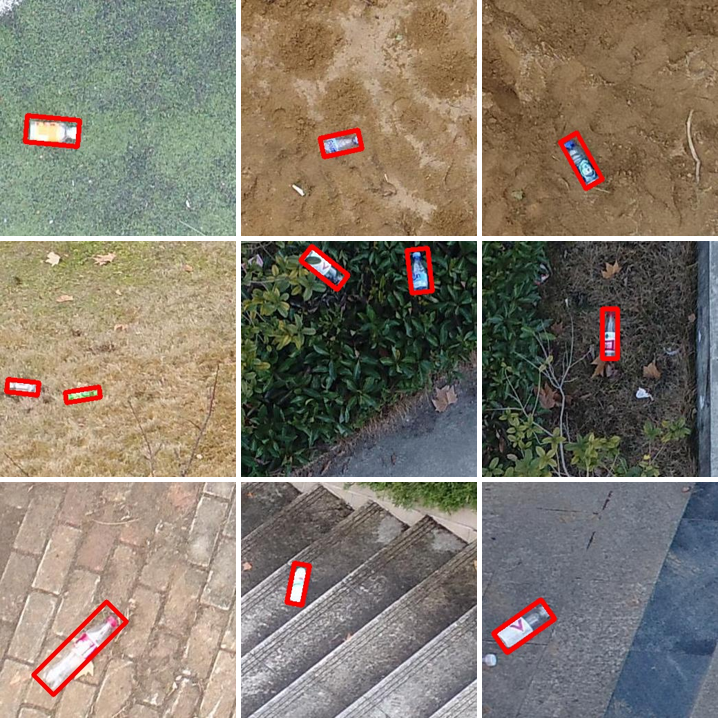}
	}
	\subfigure[ ]
	{
		\label{fig:multidet}
		\includegraphics[width=0.28\linewidth]{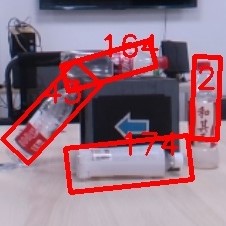}
	}
	\subfigure[ ]
	{
		\label{fig:task}
		\includegraphics[width=0.28\linewidth]{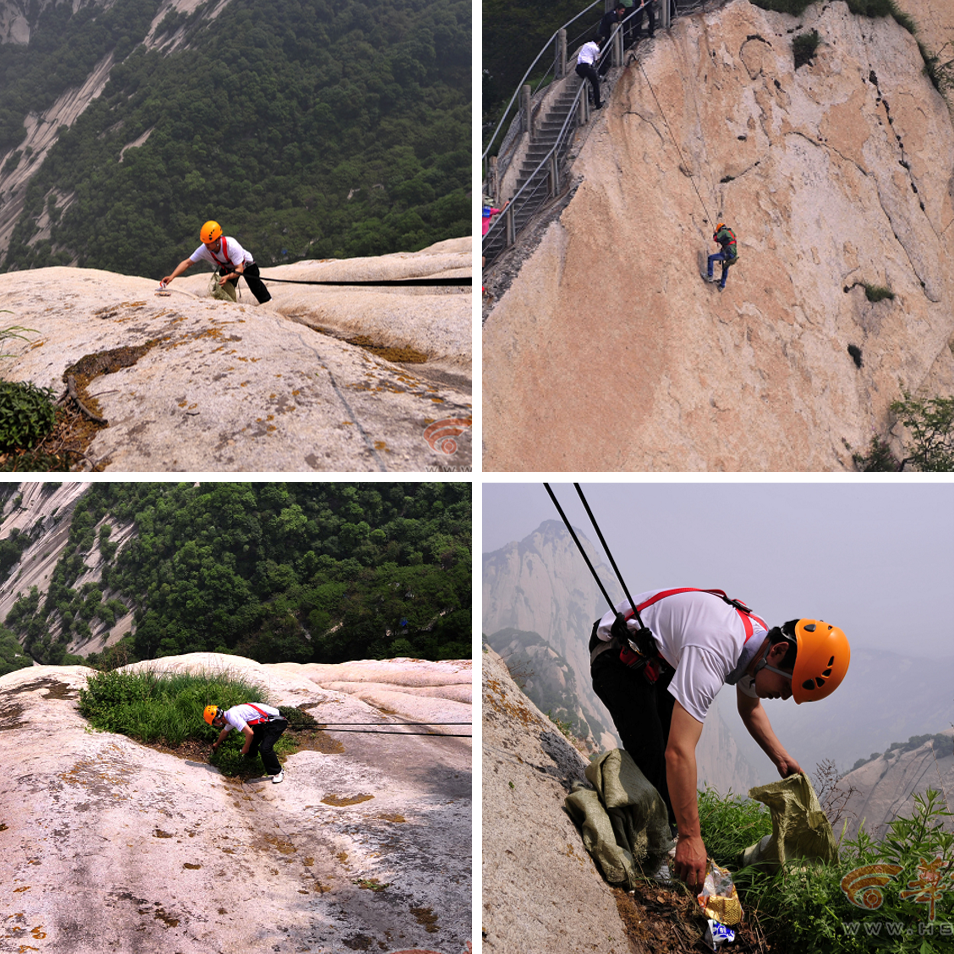}
	}
	\caption{(a) An unmanned aerial manipulator is grasping a plastic bottle to accomplish autonomous grasping tasks.
	(b) Sample images in UAV-DB dataset with plastic bottle groundtruth labeled by red box.
	(c) Detection results of Rotation-SqueezeDet with 2D box and rotation angle.
	(d) Workers are carrying out dangerous garbage disposal work on the cliff.
	}
	\label{fig:first_image}
\end{figure}

Unmanned aerial manipulators (UAMs) are known as one specific type of unmanned aerial vehicles (UAVs) equipped with one or multiple robotic arms and have attracted a lot research in recent years~\cite{review}.  One main advantage of UAM is that it shows promising potentials to transform passive sensing missions into active 3-dimension (3D) interactive missions like grasping~\cite{2dof2013} and assembling~\cite{insert2014}. 
The capabilities like aerial maneuvering and hovering make it possible for UAM to accomplish dangerous missions like grasping the rubbish on the cliff in scenic spots. 
Fig. \ref{fig:task} illustrates some dangerous and costly rubbish cleaning works.

For the autonomous aerial manipulating missions, building a controllable system is always the first step. 
However, many factors can influence the stability of the overall system, 
such as the change of Center of Gravity (CoG) generated by the movements of the robotic arm, the reaction force produced by robotic arm, the complex aerodynamics effects.
Many efforts have been done to reduce these effects. 

A comprehensive dynamic model of hexacopter with a robotic arm has been built in \cite{bovzek2017navigation}, analyzed the effects of CoG change and mass distributions.
% \cite{arleo2013control} gives the kinematics and dynamics models of quadrotor with n Degree of Freedom (DoF) robotic arm. 
For successfully accomplishing the insertion task, a two-stage cascaded PID controller has been proposed in \cite{insert2014}.
A Variable Parameter Integral Backstepping (VPIB) controller proposed in \cite{jimenez2013control} can control an UAM with better performance than PID controller.
% Based on \cite{jimenez2013control}, \cite{heredia2014control} proposed a better nonlinear controller with a consideration of full dynamics model and used an admittance controller to control the robotic arm.
With a movable compensation mechanism, a multilayer architecture controller compensate the internal and external effects layer by layer to control the UAM was presented \cite{ruggiero2015multilayer}.
To suppress the torque generated by the movements of robotic arm, a novel mechanism with a simplified model has been adopted in \cite{jpn}.

% \footnotetext{\url{http://news.hsw.cn/system/2012/05/16/051321760.shtml}}

Usually, human operators are unable to accurately control the UAM due to observation change and data transmission delay.
Such inaccuracy causes the robotic arm to be difficult to align and grasp the targets.
Therefore it is better that the UAM can work without relying on external control.
For fully autonomous grasping, the UAM needs perception ability.
However, currently only a few works\cite{visionguided}, \cite{ramon2017detection}, \cite{onvision} considered robust visual perception aided autonomous grasping.
An Image-Based Visual Servo (IBVS) was implemented in \cite{visionguided} to help locating the position of the targeted object.
Feature models are used in \cite{ramon2017detection} to find the position of known targets.
In \cite{visionguided}, correlation filters were adopted to track targets.

In recent years, End-to-End object detection algorithms like \cite{FasterRCNN}, \cite{yolov3} and \cite{SqueezeNet} have shown surprising results. 
However, attempts directly applying these methods in the UAM system usually end with failure or unsatisfactory robustness.
The reason lies in the objects, like bottles, in UAV perspective often appeared with arbitrary poses as shown in Fig. \ref{fig:UAV-BD-Samples}, since the camera is tight-coupled and rotated with the UAV body.
Moreover, most grasping missions need target's rotation angle for not touching the target when the end-effector is approaching.
Furthermore, the performance like robustness and efficiency of the detection algorithms play an important role in grasping missions since it can bring more mobility to the UAM.
However, even with NVIDIA Jetson TX2, current common oriented bounding boxes based methods like \cite{RRPN, R-DFPN} are still unable to run onboard since both require more than 11G GPU memory and Jetson TX2 only has 8G GPU memory.

% So we proposed Rotation-SqueezeDet, combine the rotation descriptor $(c_x, c_y, w, h, \theta)$ with SqueezeDet\cite{SqueezeDet}. 
% Our method 

In order to solve the problems mentioned above, we propose Rotation-SqueezeDet which can regress rotation angle and position in 2-dimension (2D) image in near real-time. 
Unlike the common horizontal bounding box descriptor $(c_x, c_y, w, h)$, where $(c_x, c_y)$ is the center location, $w$ and $h$ are the width and height of the bounding box, respectively, Rotation-SqueezeDet introduces a new $\theta$ term, and thus uses $(c_x, c_y, w, h, \theta)$ as descriptor to describe object position and rotation angle in the 2D image.
This not only makes the detection more robust since the bounding box included fewer background, but also provides the rotation angle of the target.
By using Intel RealSense D435 depth camera, the relative 3D distance of targets can be measured in the point cloud generated by registered depth image once the target is detected. 
Hence, a rotation-aware grasping for autonomous grasping is possible. 
A glimpse of detection results is shown in Fig. \ref{fig:multidet}.

When applying the UAM into real world scenes, the flow generated by rotors can easily blew the lightweight targets away, like empty plastic bottles, and it is hard to predict whether an object can be easily blew away by the downward flow or not.
Hence, a safety grasping action is better to be taken under weak or no flow influenced conditions.
High-fidelity Computational Fluid Dynamics (CFD) simulation results of many different UAVs have been presented in \cite{NASA}.
From the observation of these results, the flow generated by each UAV's rotor is decreasing rapidly in the outter-wing area. 
Further, we roughly confirmed this observations from experiments by using an digital anemometer. 
To reduce the flow effects, the design and motion planning of robotic arm are in the light of the CFD sumulations and our flow measurements.
Our experimental results have shown that grasping under weak or no flow influenced is possible.

The main contributions of this letter can be summarized as follows:
\begin{enumerate}
	\item The Rotation-SqueezeDet method is proposed. This method can run on Jetson TX2 in near real-time and enable successful rotation-aware grasping. 
	% To the best of our knowledge, it's the first time the rotation-detection methods is applied in UAM for autonomous grasping mission.
	We believe that this method will be suitable in not only the aerial grasping but also general missions. 	
	\item We consider the flow influence for designing and moving the robotic arm. The designed UAM can grasp lightweight objects in weak or no flow influenced area during flight.
	\item We have designed and assembled the UAM platform. The system is affordable since it costs less than \$2300 USD and can fly without relying on expensive visual motion caption system. 
	All mechanical structures are provided as open--hardware for reuse by the community. Link:\url{https://github.com/eleboss/UAMmech}
\end{enumerate}

The rest of this letter is organized as follows.
Section \ref{section:system} describes the design and details of the overall system. 
Section \ref{section:arm} describes the motion planning and control of the robotic arm. 
Section \ref{section:perception} describes the complete vision system including Rotation-SqueezeDet.
Section \ref{section:experiments} experimentally demonstrates the system including autonomous grasping and vision system performance.
Finally, the conclusion and future work are presented in Section \ref{section:conclusion}.

% % Start with the problem mentioned above, with proposed 

% In this work, we focus on the on-board perception system for autonomous object grasping.

% In computer vision, many visual concepts such as object are described with bounding boxes. 

% Objects without many orientations can be described with this method. 
% However, some objects such as bottles in UAV images often appear with arbitrary orientations depending on the shooting angle of the UAV camera. 
% In the UAV arm grabbing task, object's direction is very important, it can help claw to grab object with true angle. 
% So, it's necessary to use $(c_x, c_y, w, h, \theta)$ to describe object position and direction.

% For a prospective and valuable detection method in our system, there are two essentials that authentically matter: one is an available implementation on the platform of low power consumption, the other is a high frame rate to meet the requirements of real time processing. 
% We choose NVIDIA Jetson TX2 as the core of our embedded system, which balanced the power-efficient and the computing power. 
% At the same time, it's very important for grasping task use UAV to run in real time. There are some $\theta-\rm{based}$ object detection algorithms such as \cite{RRPN, R-DFPN} which are based on Faster R-CNN, they can't be runned on NVIDIA Jetson TX2. 
% So we proposed a method called Rotation-SqueezeDet, which is based on SqueezeDet. It can be runned on TX2 with about 17FPS.

\section{System Description}
\label{section:system}

% In this section, we first present the notation system applied in this letter, and next the planning of robotic arm and design of displacement compensation system (DCS). Finally, the controller for robotic arm and DCS is detailed.
\subsection{Notation}
\begin{figure}
    \centering
	\includegraphics[width=0.95\linewidth]{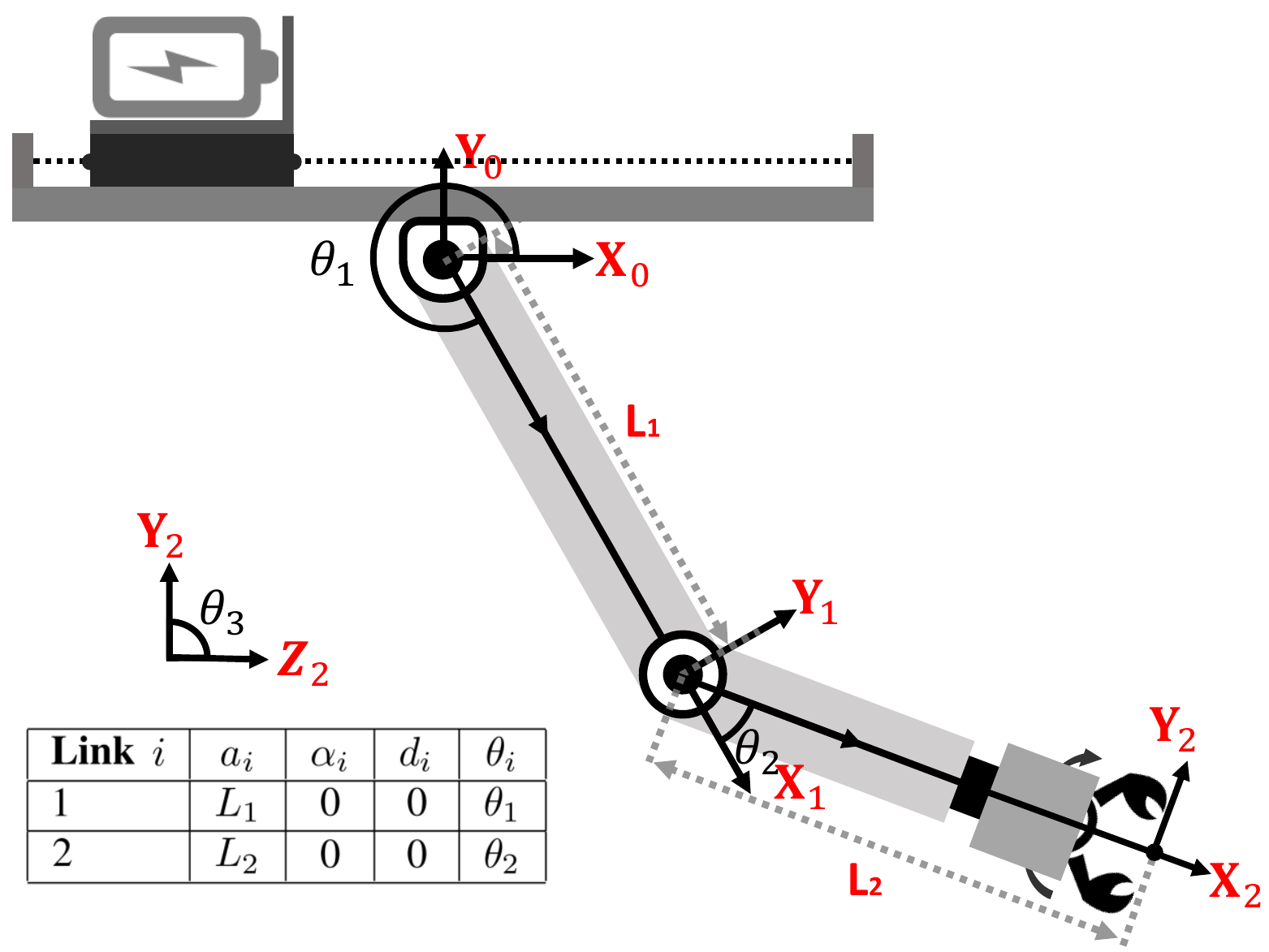}
	\caption{Coornidates system and D-H parameter of the robotic arm.}
	\label{fig:arm}
\end{figure}

The East, North and Up (ENU) coordinate system is used as world-fixed inertial frame corresponding to $\{x_w,y_w,z_w\}$. 
Following the definition of well known Denavit-Hartenberg (D-H) parameters \cite{robo}, the link frame of Link $i$ is defined as the $\{x_i,y_i,z_i\}$, $i=0$ is the fixed arm frame. The definition of link frame is detailed in Fig. \ref{fig:arm} and the table in the bottom left gives the D-H parameters of robotic arm.
The body frame is assumed to be the geometrical center of the UAM denoted as $\{x_b,y_b,z_b\}$. 
$G_x$, $G_y$, $G_z$ indicate the CoG of the UAM in body frame $\{x_b,y_b,z_b\}$.
% And the origin of the body frame is attached to arm fixed axis origin for simplicity.
$(\phi, \theta, \psi)$ indicate the roll-pitch-yaw Euler angels. 
$\{x_t,y_t,z_t\}$ define the detected targets in the RealSense D435 camera frame.

% Transformations between frames can be done by rotation matrix R $\in$ SO(3), and defined as $B^R_W$. 

\begin{figure}
    \centering
	\includegraphics[width=0.8\linewidth]{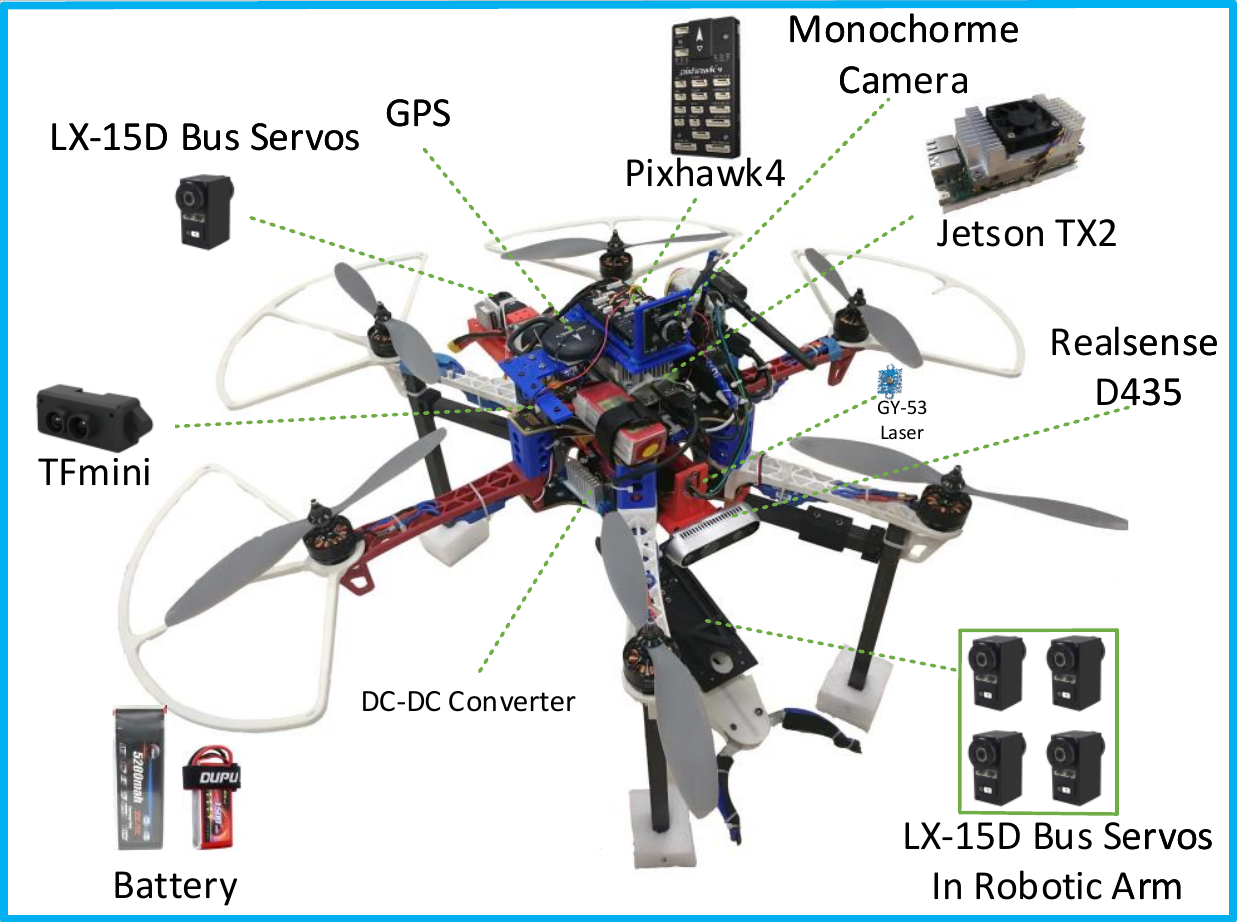}
	\caption{Hardware components of the UAM}
	\label{fig:hardware}
\end{figure}

\subsection{Hardware}
In this work, a modified DJI hexacopter frame F550\footnote[1]{\url{www.dji.com/flame-wheel-arf/feature}} is adopted. 
The hardware components are shown in Fig. \ref{fig:hardware}. 

The propulsion module of the UAM is composed of Sunnysky\footnote[2]{\url{www.rcsunnysky.com}} x3108s motor, Hobbywing\footnote[3]{\url{www.hobbywing.com}} platinum Electronic Speed Controller (ESC), and 10 inches propeller. 
The Pixhawk4\footnote[4]{\url{www.holybro.com/product/55}} autopilot with PX4\footnote[5]{\url{https://github.com/PX4/Firmware}} V1.8.0 flight stack and NVIDIA Jetson TX2\footnote[6]{\url{developer.nvidia.com/embedded/buy/jetson-tx2}} are placed at the top of the UAM as the main computing devices. 
A global shutter monochrome camera\footnote[7]{\url{www.jinyandianzi.com}} is tight-coupled with Pixhawk4 by using a 3D-printed anti-vibration damping plate.
A Benewake TFmini\footnote[8]{\url{www.benewake.com/tfmini.html}} laser rangefinder is chosen for being cheap, lightweight and with up to 12m maximum detection range, mounted downward facing to privide altitude feedback.
An Intel RealSense D435\footnote[9]{\url{www.realsense.intel.com/stereo}} camera is mounted at the middle of the drone facing forward to find the targets. 

We build a displacement compensation system  (DCS) which can move counterweight to align the CoG and thus improves the stability of the total system. The DCS is mounted in the middle of UAM and made by 3D printed PLA material, including tow rails, a slide table and a bus servo to provide drive force.
The LEBOT\footnote[10]{\url{www.lobot-robot.com}} LX-15D serial bus servos are chosen for being budget friendly, lightweight, and having multiple extra structures to facilitate the installation.
Another significant advantage of the bus servo is it can largely reduce the wiring complexity and control difficulty. 
So we can only use one serial port in Jetson TX2 to control all servos. 
While the LX-15D servo can only provide $240^\circ$ feedback, we use a short range time of flight (ToF) laser rangefinder GY-53 to provide the battery position feedback.

A 5200mAh 4S-35C battery weighted 0.525kg is used for providing enough power to the propulsion system and as counterweight for the DCS. And this battery can sustain the flight time around 3min.
Another 1500mAh 3S-30C weight 0.138kg battery is used for providing power to the robotic arm and computing facilitates. 

\begin{figure}
    \centering
	\includegraphics[width=\linewidth]{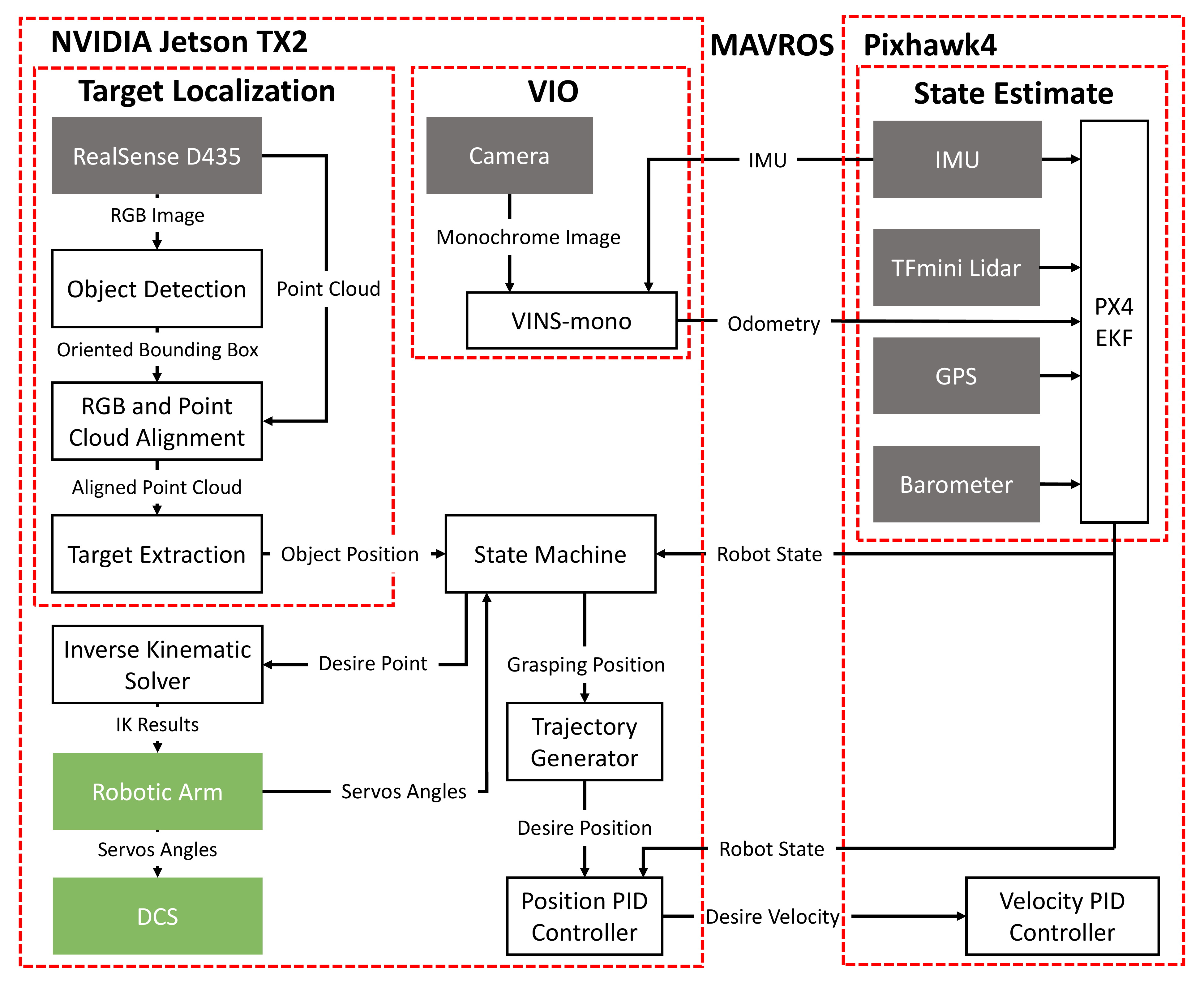}
	\caption{An overview of the software architecture.}
	\label{fig:software}
\end{figure}

% \subsection{Robotic Arm and DCS}
Drive forces of robotic arm are also provided by LX-15D servos. 
The robotic arm has 3-DoF and can grasp objects by the end-effector. 
The first 2-DoF provide the robotic arm mobility to move at the planar 2D plane. 
The last DoF enables a rotational grasping by cooperating with vision system. 
The last DoF is of vital important for the rotated objects grasping, since the unrotated grasping can easily tip the object. 

The total takeoff weight of the UAM is about 4.08kg. 
Thanks to the carbon fibre material and Poly Lactic Acid (PLA) material, the robotic arm is only weighted 0.459kg and has 43cm extended range.

\subsection{Software Architecture}
The software runs on two main processors: Pixhawk4 and Jetson TX2, and
all processes are running onboard. 
Fig. \ref{fig:software} gives an overview of the system software architecture.
Note that the Jetson TX2 is overclocked to run at max clock rate and unlocked 2 external CPU cores to generate more computing power.

Robot Operating System (ROS)\cite{ros} is a pseudo-operating system which allows developers to work cooperating by following its running mechanism.
Modules related to state estimation and flying control are running on Pixhawk4 and exchange data with Jetson TX2 by MAVROS\footnote[11]{\url{https://github.com/mavlink/mavros}}.
Visual Inertial Odometry (VIO) and object detection algorithms are running on Jetson TX2. 
A state machine is adopted to set the robotic arm motion and UAM waypoint. 
% The flying trajectory is generated by a simple trajectory generator which will let the UAM align to the objects grasped point in the order of XYZ axis.
So the grasping position $(x_w^s, y_w^s, z_w^s)$ of the UAM can be given by:
\begin{equation}
    \begin{aligned}
		\left[
			\begin{matrix}
				x_w^s \\
				y_w^s \\
				z_w^s \\
				1
			\end{matrix}
		\right]
		= 
		\left[
			\begin{matrix}
				x_w \\
				y_w \\
				z_w \\
				1
			\end{matrix}
		\right]
		+ 
		T_B^W T_C^B 
		\left[
			\begin{matrix}
				x_t \\
				y_t \\ 
				z_t \\
				1
			\end{matrix}
		\right]
		- 
		T_B^W T_0^B
		\left[
			\begin{matrix}
				x_0^g \\ 
				y_0^g \\ 
				0 \\
				1
			\end{matrix}
		\right]
	\end{aligned}
\end{equation}
where $T_B^W, T_0^B, T_C^B \in \mathbb{R}^{4 \times 4}$ are the homogeneous transformation matrix, 
$T_0^B$ transforms the fixed arm frame to body frame, 
$T_C^B$ transforms the camera frame to the body frame,
$T_B^W$ transforms the body frame to the world-fixed frame,
$(x_0^g, y_0^g)$ is the grasping point.

We use the cascaded PID controller to control the drone, the position loop runs at TX2, the velocity loop runs at the Pixhawk4.
The parameters of the cascaded PID are tuned when the robotic arm keeps static at the yellow star point $(x_0^f, y_0^f)$ shown in Fig. \ref{fig:space}.

\subsection{State Estimation}
State estimation is the foundation of our system as it provides crucial information to help other parts to achieve the best performance.
In this work, we use the Pixhawk4 built-in Extended Kalman Filter (EKF) to fuse multiple sensors feedback for state estimation. 

Global Positioning System (GPS) usually fails to provide global positioning feedback when in indoor environment.
Hence, in order to fly indoor, without using expensive visual motion caption system, we integrate the VINS-mono\cite{vins} VIO to provide the local position feedback, and it can provide the highest level of accuracy and robustness compared with multiple VIO\cite{slambenchmark}.
The VINS-mono runs at 10hz with loop-closure and the output is rotated to ENU world-fixed frame denoted $(x_w^v, y_w^v, z_w^v)$.
For the VINS-mono, we use the Inertial Measurement Unit (IMU) in Pixhawk4 as the inertial input, and a monochrome global shutter camera runs at $640\times 400$ resolution and 90 Frames Pre Second (FPS) to provide clear images as visual input.
To reduce drifting, the monochrome camera and Pixhawk4 are tight-coupled by using the 3D printed structures, the camera intrinsic matrix and camera to imu transformation parameters are carefully calibrated by using kalibr\cite{kalibr}.
% Further, the accelerometer and gyroscope measurement noise standard deviation parameters are increased to 0.4 and 0.04 in the config files to resist the vibration noise.
Due to the computation limitation and data transmission delay, the output of VINS-mono runs in Jetson TX2 has about 140ms delay compared with current IMU output. 
In order to synchronize these outputs, we first calculate the velocity of VINS-mono estimation $(\dot{x}_w^v, \dot{y}_w^v, \dot{z}_w^v)$, then apply some random movements to the UAM, so the delay time can be found by comparing $(\dot{x}_w^v, \dot{y}_w^v, \dot{z}_w^v)$ with the Pixhawk4 velocity estimation.
Finally, $(x_w^v, y_w^v, z_w^v)$ is used as external vision aid of the Pixhawk4 onboard EKF to give 100hz state estimation.

For robust flying, the main altitude feedback is not given by the VINS-mono but the TFmini rangefinder. 
The total delay of TFmini measurement is about 30ms.
\section{ROBOTIC ARM \& DCS}
\label{section:arm}

\subsection{Robotic Arm Motion Planning}
In order to find the workspace of robotic arm, the forward kinematics of a 3DoF robotic arm is given by the following equations:
\begin{align}
    & x_0 = L_1\cos\theta_1 + L_2\cos(\theta_1+\theta_2)\nonumber \\ 
    & y_0 = L_1\sin\theta_1 + L_2\sin(\theta_1+\theta_2)\\
    & \theta_3 = \theta \nonumber
\end{align}
where $L_1$, $L_2$ are the lengths of the first and the second links. 
$\theta_1$, $\theta_2$, $\theta_3$ are the rotation angles of each joint, 
$(x_0, y_0)$ is the point in arm fixed frame.
Since the $\theta_3$ is only related to target rotation angle $\theta$, the workspace of our robotic arm is equal to a planar 2DoF robotic arm model. 
The table presented in Fig. \ref{fig:arm} gives a clear D-H parameters definition of the robotic arm.

\begin{figure}
    \centering
	\includegraphics[width=0.9\linewidth]{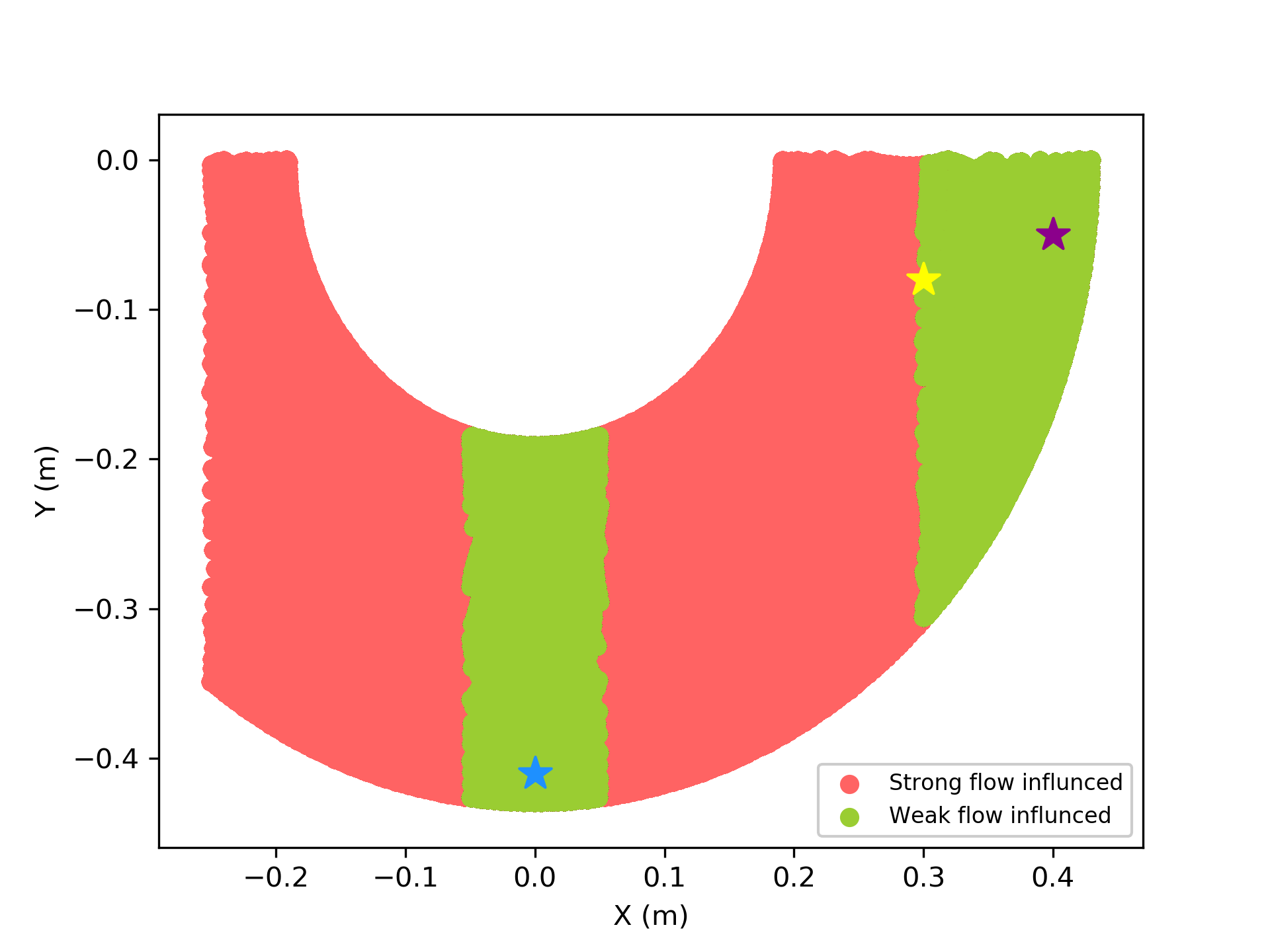}
    \caption{Workspace of robotic arm printed by using forward kinematics considered flow influence.}
	\label{fig:space}
\end{figure}

Based on the flow measurements which will be described in section \ref{sec:Flow_Distribution_Validation} and mechanical limits, the actual workspace is given in Fig. \ref{fig:space}, the green points indicate weak or no flow influenced area, red points indicate strong flow influenced area.
The blue star point is the dropping point and the purple star point is the grasping point $(x_0^g, y_0^g)$.
The robotic arm holds at the yellow star point $(x_0^f, y_0^f)$ during flight. 
And all these points can be redefined depend on the applications.
We use Inverse Kinematics (IK) to solve the desired rotation angle of each joints.
Assumed the robotic arm is planning to move to $(x_0, y_0)$ point, the IK result is given by:
\begin{eqnarray}
    \theta_2 = \pm\arccos(\frac{{x_0}^2 + {y_0}^2 - {L_1}^2 - {L_2}^2}{2L_1L_2}) 
\end{eqnarray}
Here $\theta_2 \in [0, \pi]$ is downward elbow solution, and $\theta_1$ can be derived by:
\begin{equation}
    \begin{aligned}
        \theta_1 = {} &\arctan({x_0}^2 / {y_0}^2) - \\ 
         & \arccos(\frac{{x_0}^2 + {y_0}^2 + {L_1}^2 - {L_2}^2}{2L_1\sqrt{{x_0}^2 + {y_0}^2}})
    \end{aligned}
\end{equation}

\subsection{CoG Compensation}
When the UAM is static on the ground and the robotic arm hold static at $(x_0^f, y_0^f)$, a symmetry placement design is utilized to make sure the $G_x$ and $G_y$ are fitted with the geometry center.

However, movements of the robotic arm can change the $G_x$. 
For the dynamic $G_x$ alignment, we adopte the strategy presented in \cite{ruggiero2015multilayer} called DCS, moving the battery as a counterweight since it's weight can provide sufficient compensation in relatively short moving distance.

CoG transformation of Link $i$ and the end-effector payload from link frame to the body frame is given by:
\begin{eqnarray}
    \left[
        \begin{matrix}
        x_{bi}^g \\
        y_{bi}^g \\
        z_{bi}^g \\
        1
        \end{matrix}
    \right] 
    = 
    T_0^B T_i^0 \left[
        \begin{matrix}
        x_i^c \\ 
        y_i^c \\
        z_i^c \\
        1
        \end{matrix}
    \right]
\end{eqnarray}
where $T_i^0, T_0^B \in \mathbb{R}^{4 \times 4}$ are the homogeneous transformation matrices, $T_0^B$ transforms from fixed arm frame to body frame, $T_i^0$ transforms from each link frame to the fixed arm frame.
($x_i^c$, $y_i^c$, $z_i^c$) is the CoG position of Link $i$ in the fixed arm frame.
($x_{bi}^g$, $y_{bi}^g$, $z_{bi}^g$) is the CoG position of Link $i$ in body frame, here $i = 3$ indicates the grasped object.

To align the $G_x$ at geometry center, a linear slider is designed to move the battery and the position of the battery $p_b$ in the body frame can be calculated by:
\begin{eqnarray}
    p_b=\frac{\sum_{i=1}^3m_ix_{bi}^g}{m_b}
\end{eqnarray}
where $m_i$ is the mass of Link $i$, $m_b$ is the mass of battery .

The displacement compensation plays a key role in stabilizing the UAM. Without the DCS, the change of CoG can easily make aside rotor reach the maximum thrust leading to unstable. 
And this method works well in limited speed movement of robotic arm. 
To guarantee the compensation performance, we limit the rotation speed of each joint in robotic arm to make sure it does not exceed the maximum compensation speed of the linear slider.

\subsection{Control}
\begin{figure}
    \centering
	\includegraphics[width=\linewidth]{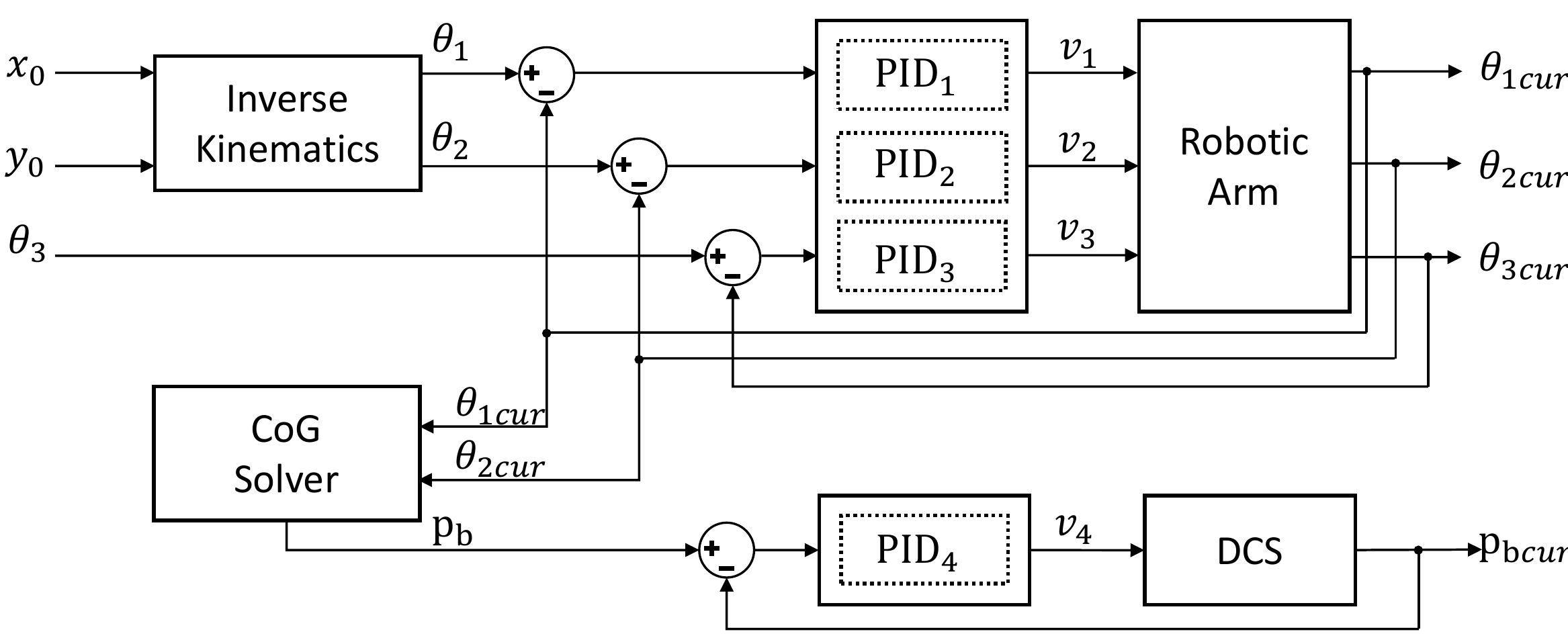}
	\caption{Control diagram of the DCS and robotic arm.}
	\label{fig:pid}
\end{figure}

The total control diagram of the robotic arm and DCS are shown in Fig. \ref{fig:pid}. In Fig .\ref{fig:pid}, $v_1, v_2, v_3, v_4$ is the rotation speed of the servos and the $\theta_{1cur}, \theta_{2cur}, \theta_{3cur} , p_{bcur}$ indicate the current feedback.
Three linear PID controllers are adopted to control the robotic arm.
Another PID controller uses the position of the battery $p_b$, detected by a laser rangefinder, to control the linear slider.
% For the feedback of $p_b$, we are not directly transform the angle feedback to distance because it is not accurate due to the limited servo angle feedback and mechanical deviation.
All PID parameters are well tuned to guarantee a stable and smooth control.

\section{VISION SYSTEM}
\label{section:perception}
In this section, we introduce the vision system including a light and fast oriented-object detection model called Rotation-SqueezeDet and a target localization framework based on Rotation-SqueezeDet detection results and point clouds from the depth camera.

Our proposed model is inspired by SqueezeDet\cite{SqueezeDet} and Rotation Region Proposal Networks (RRPN)\cite{RRPN}. As the former, SqueezeDet is a single-pass detection pipeline combining bounding box localization and classification by a single network. It appears to be the smallest object detector by virtue of powerful but small backend network of SqueezeNet\cite{SqueezeNet, LCDet}. As the latter, RRPN is based on Faster R-CNN\cite{FasterRCNN}, it can detect oriented objects. The difference between Faster R-CNN and RRPN is described as below. In Faster R-CNN, the Region of Interests (RoIs) are generated by Region Proposal Network (RPN), and the RoIs are rectangles which can be written as $R = (x_{min}, y_{min}, x_{max}, y_{max}) = (c_x, c_y, w, h)$. These RoIs have regressed from $k$ anchors which are generated by some predefined scales and aspect ratios. However, in RRPN, instead, it uses Rotation anchors (R-anchors) and rotation RoI pooling, brings ability to predict oriented bounding boxes denoted as $R = (c_x, c_y, w, h, \theta)$.

Object detection algorithms like Faster R-CNN have high accuracy but slow processing speed and large storage requirement. Moreover, RRPN is slower than Faster R-CNN, it takes twice as much time as the Faster-RCNN\cite{RRPN}. Thus RRPN does not meet our run-time requirement. Considering a comparable accuracy and run-time on Jetson TX2, SqueezeDet is a suitable choice, it can run about 45FPS with $424\times 240$ pixels image on Jetson TX2 and easy to train. However, SqueezeDet cannot predict the $\theta$ of rotated object because it uses horizontal bounding boxes. So we designed a model which can generate oriented bounding boxes based on SqueezeDet and named Rotation-SqueezeDet, it can run on Jetson TX2 in near real time.

\begin{figure}
    \centering
	\includegraphics[width=\linewidth]{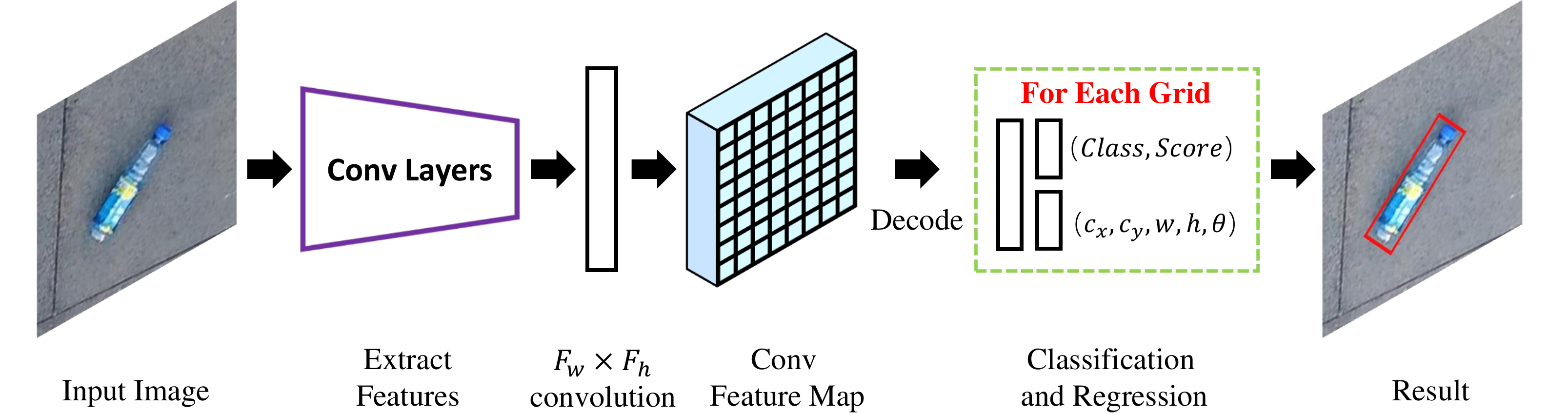}
	\caption{Rotation-SqueezeDet's detection pipeline.}
	\label{fig:pipeline}
\end{figure}
\subsection{Network Architecture}
The overall model of Rotation-SqueezeDet is illustrated in Fig. \ref{fig:pipeline}. In this model, a convolutional neural network, SqueezeNet V1.1, first takes an image as input and extracts a low-resolution, high dimensional feature map from the image. Then the feature map is fed into the $F_w\times F_h$ convolutional layers to compute oriented bounding boxes at each position of conv feature map. Next, each oriented bounding box is associated with $(5+C+1)$ values, where $5$ is the number of bounding box parameters, $C$ is the number of classes, and $1$ is the confidence score. And each position on the conv feature map computes $K\times (5+1+C)$ values that encode the bounding box predictions. Here, $K$ is the number of R-anchors, each R-anchor can be described by $5$ parameters as $(c_x^a, c_y^a, w^a, h^a, \theta^a)$, $(c_x^a, c_y^a)$ are R-anchor's center on image, $w^a, h^a, \theta^a$ are the width, height and angle of R-anchor, respectively.
Notice that $(c_x, c_y, w, h, \theta)$ are decoded from predicted parameters tuple $\bm{v}$, which is the direct outputs from the conv feature map. Here $\bm{v}$ are encoded as follow:
\begin{align}
& v_x = \frac{c_x - c_x^a}{w^a}, v_y = \frac{c_y - c_y^a}{h^a}, \nonumber\\
& v_w = \log\frac{w}{w^a}, v_h = \log\frac{h}{h^a}, v_{\theta} = \theta - \theta^a + k\pi
\end{align}
where $(c_x, c_y, w, h, \theta)$ are parameters describe the predicted oriented bounding box, $(c_x^a, c_y^a, w^a, h^a, \theta^a)$ are parameters describe the R-anchor, and here $k\in Z$ to ensure $\theta \in [0, \pi)$.

\subsection{Oriented IoU Computation}
To efficiently predict the rotation angle of object, we introduce a parallel IoU computing method. First, we attempt to calculate the IoU using the OpenCV's functions \textit{rotatedRectangleIntersection} and \textit{contourArea} directly. However, the efficiency of these functions are poor because they cannot compute parallelly. Thus, we use a simple and efficient method to approximately compute the IoU parallelly, which is to use the angle deviation of two oriented bounding boxes. The approximate IoU\cite{ADPF} can be computed by:
\begin{eqnarray}
    \rm{IoU}^{\star} = \rm{IoU} * \rm{abs}(1 - \frac{\theta_1^b - \theta_2^b}{\pi})
\end{eqnarray}
where $\theta_1^b$ and $\theta_2^b$ are the rotation angle of two oriented bounding boxes, IoU is computed by treating oriented bounding boxes as horizontal bounding boxes.

\subsection{R-Anchors' Selection}
The R-Anchors are different from the horizontal anchors. And for the R-Anchors' selection, we use a K-means based method described in \cite{YOLOv2} to select R-anchors' $w$ and $h$ to match the data distribution, we set $k$ as $9$ in K-means and treat objects' angle distribution as a uniform distribution, i.e., we set R-anchors' angle as $\{0, \frac{\pi}{9}, \frac{2\pi}{9}, \frac{3\pi}{9}, \frac{4\pi}{9}, \frac{5\pi}{9}, \frac{6\pi}{9}, \frac{7\pi}{9}, \frac{8\pi}{9}\}$. Therefore there are $81$ anchors at each conv feature map position.

\subsection{Object Localization}
To acquire the real world position of the target, we use the RGB-D camera to detect and locate the target. First, the aligned RGB image and point clouds can be obtained by aligning RGB image and depth image in the same coordinate system. Next, the subarea of the total point clouds containing location information of the target which can be extracted from the whole point clouds by utilizing the detection result $(c_x, c_y, w, h, \theta)$. After that, we use a small central subarea of target's point clouds to calculate its real world position $(x_t, y_t, z_t)$ in the camera frame given by:
\begin{eqnarray}
    (x_t, y_t, z_t) = (\frac{1}{L}\sum_{i=0}^{k^2} \mathbf{X}_p^i, \frac{1}{M}\sum_{i=0}^{k^2} \mathbf{Y}_p^i, \frac{1}{N}\sum_{i=0}^{k^2} \mathbf{Z}_p^i)
\end{eqnarray}
where $(\mathbf{X}_p, \mathbf{Y}_p, \mathbf{Z}_p)$ are the coordinates of the point clouds set of the center subarea of target's bounding box, its superscript indicates the $i^{th}$ point value started from top left corner. $L, M, N$ are the valid points number of $\mathbf{X}_p, \mathbf{Y}_p, \mathbf{Z}_p$, respectively. The central subarea of target's point clouds can be written as $(c_x, c_y, k, k)$, the $k\times k$ is the size of selected central subarea of target's point clouds, which can be calculated by:
\begin{displaymath}
	k = \left\{
	\begin{array}{ll}
		5 & \textrm{if $\min(w, h) > 5$} \\
		\min(w, h) & \textrm{otherwise} \\
	\end{array} \right.
\end{displaymath}
Here we set $k$ as $5$ to reduce the computing burden. Finally, the parameters of position and rotation angle of target can be written as: $(x_t, y_t, z_t, \theta)$, $\theta$ is the rotation angle of target's anchor.

\section{EXPERIMENTS}
\label{section:experiments}
% We experimentally verified our system by conducting lots of actual flight tests.
The experimental setup is mentioned in Section \ref{section:system}.
During the flight tests, all data are logged onboard with no external data transmission.
A high resolution video about the experiments are available here:\url{http://youtu.be/v_hGzN8VIAU}
% \begin{figure}
%     \centering
% 	\includegraphics[width=0.7\linewidth]{images/bottle_UAV.png}
% 	\caption{Sample of annotated images in UAV-BD.}
% 	\label{fig:UAV-BD-Samples}
% \end{figure}

\subsection{Vision System Results}

We trained and evaluated our vision system based on our pervious work UAV-BD\cite{UAV-BD}, a bottle image dataset under UAV perspective. 
It contains about $34,791$ object instances in $25,407$ images labelled by oriented bounding boxes. 
For training and evaluating our model, $64\%$ of the images were randomly selected as the training data, $16\%$ as validation data and the rest $20\%$ as the testing data.

All object detection experiments were implemented on TensorFlow\cite{abadi2016tensorflow}. 
We used the pertrained model, SqueezeNet v1.1, to initialize the network. 
And the system was trained $100\rm{k}$ steps with a batch size of $20$ and a learning rate of $0.01$. Besides, weight decay and momentum were $0.0001$ and $0.9$, respectively. The optimizer was \textit{MomentumOptimizer}.

\begin{figure}
	\centering

	\subfigure[ ]
	{
		\label{fig:PR-Curve}
		\includegraphics[width=0.43\linewidth]{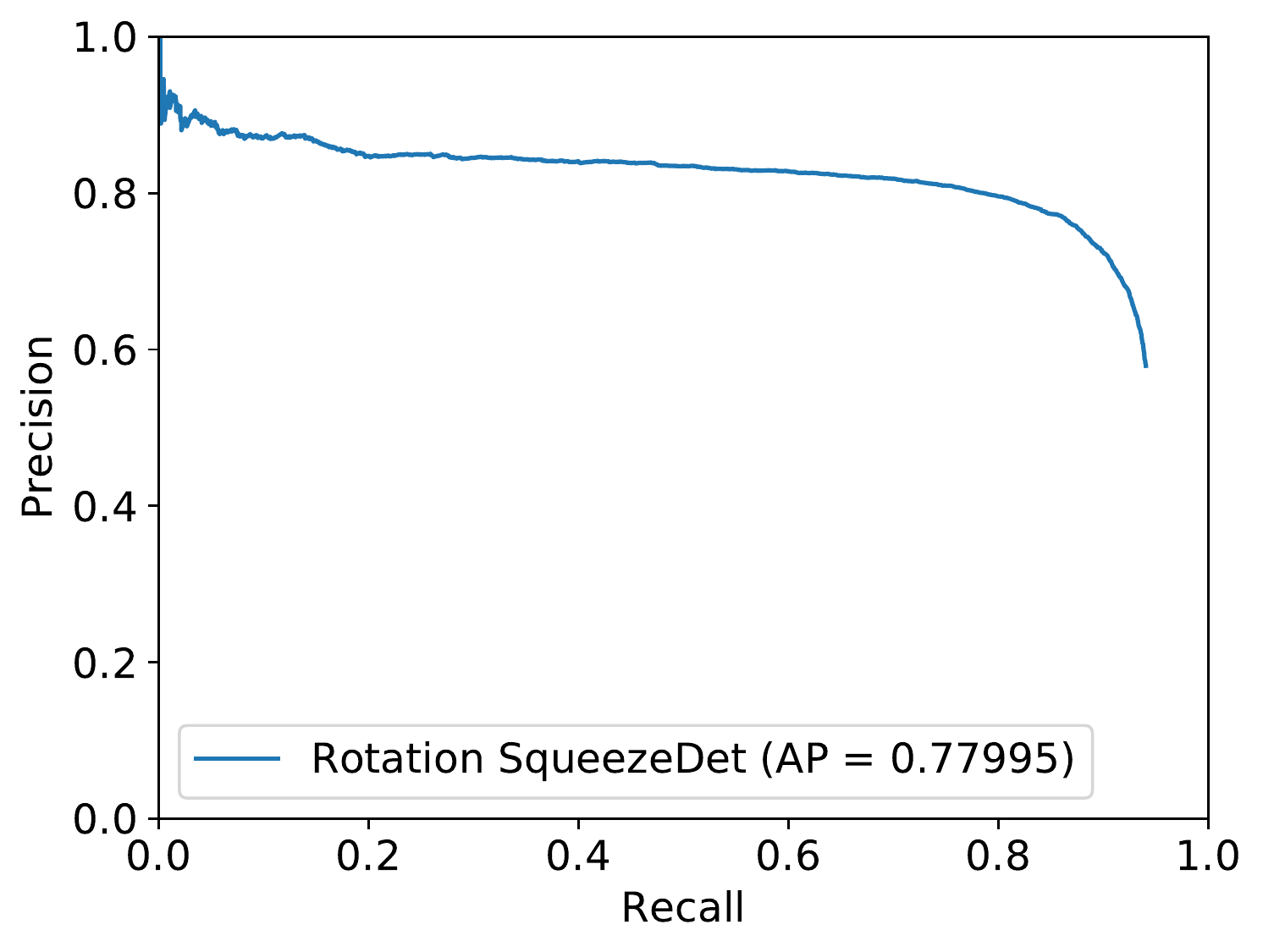}
	}
	\subfigure[ ]
	{
		\label{fig:wind}
		\includegraphics[width=0.475\linewidth]{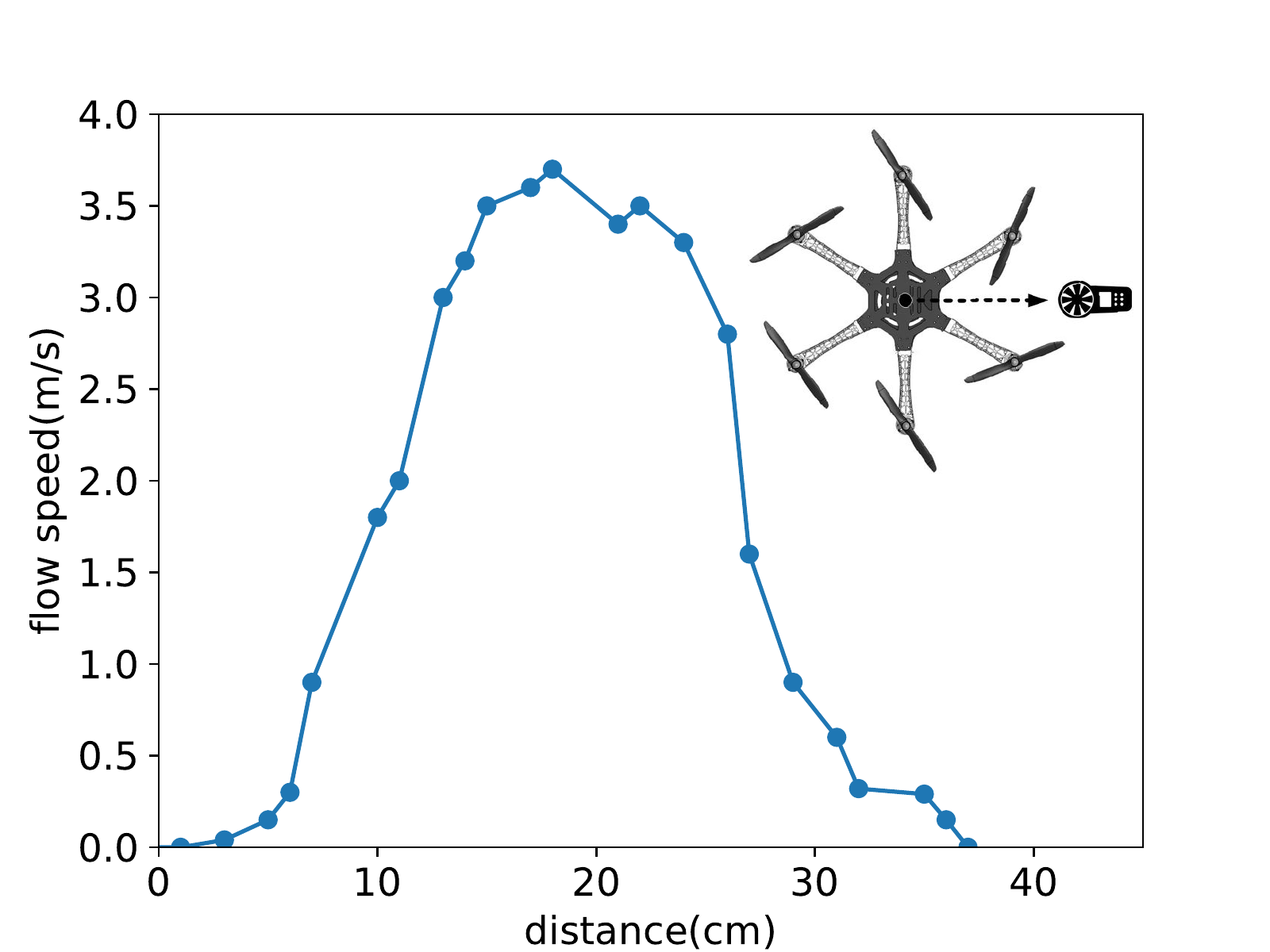}
	}
	\caption{(a) AP results of Rotation-SqueezeNet validated in UAV-BD.
	(b) The flow speed measured by digital anemometer and the subfigure in the top right corner is the moving path of the anemometer during measurement.
	}
	\label{fig:first_image}
\end{figure}

As shown in Fig. \ref{fig:PR-Curve}, the AP of Rotation-SqueezeDet on UAV-BD is about $78.0\%$ when $\rm{IoU}=0.5$. The run-time on Jetson TX2 is about $41 \rm{ms}$ with the size of $424\times240$ pixels color image. 

\subsection{Flow Distribution Validation}
\label{sec:Flow_Distribution_Validation}
In order to verify the simulation results presented in \cite{NASA} and give a rough parameter estimation for the planning mentioned in Section \ref{section:arm}.
We disarmed the UAM on the ground and used Smart AS856\footnote[12]{\url{en.smartsensor.cn/products_detail/productId=248.html}} anemometer to measure the downward flow distribution.
Since the flow is highly complex, we cannot give a very accurate flow distribution by just using an anemometer, thus we only measured a 2D flow distribution to give a rough estimation for the robotic arm planning. 
The moving path of anemometer is given in the top right corner of Fig. \ref{fig:wind} and the vertical distance from the anemometer to the rotor is about 16cm.
Following this path, we recorded the average value of $5$ measurements at one point and drew the curve in Fig. \ref{fig:wind}. Hence, the horizontal axis is the distance from the central position of UAM body to the outside following this path. 
From the observation of Fig. \ref{fig:wind}, the flow speed is decreasing rapidly at about 8cm and 30cm, and reaching the top speed at about 21cm which is the underside of rotors.
The 8cm is close to inside of the UAM, the 30cm is close to the outside of the UAM. So the flow speed is relative weak when at the central position and outside position of the UAM, this observation basically agrees with the simulation results in \cite{NASA}.

\begin{figure}
	\centering
	\includegraphics[width=0.9\linewidth]{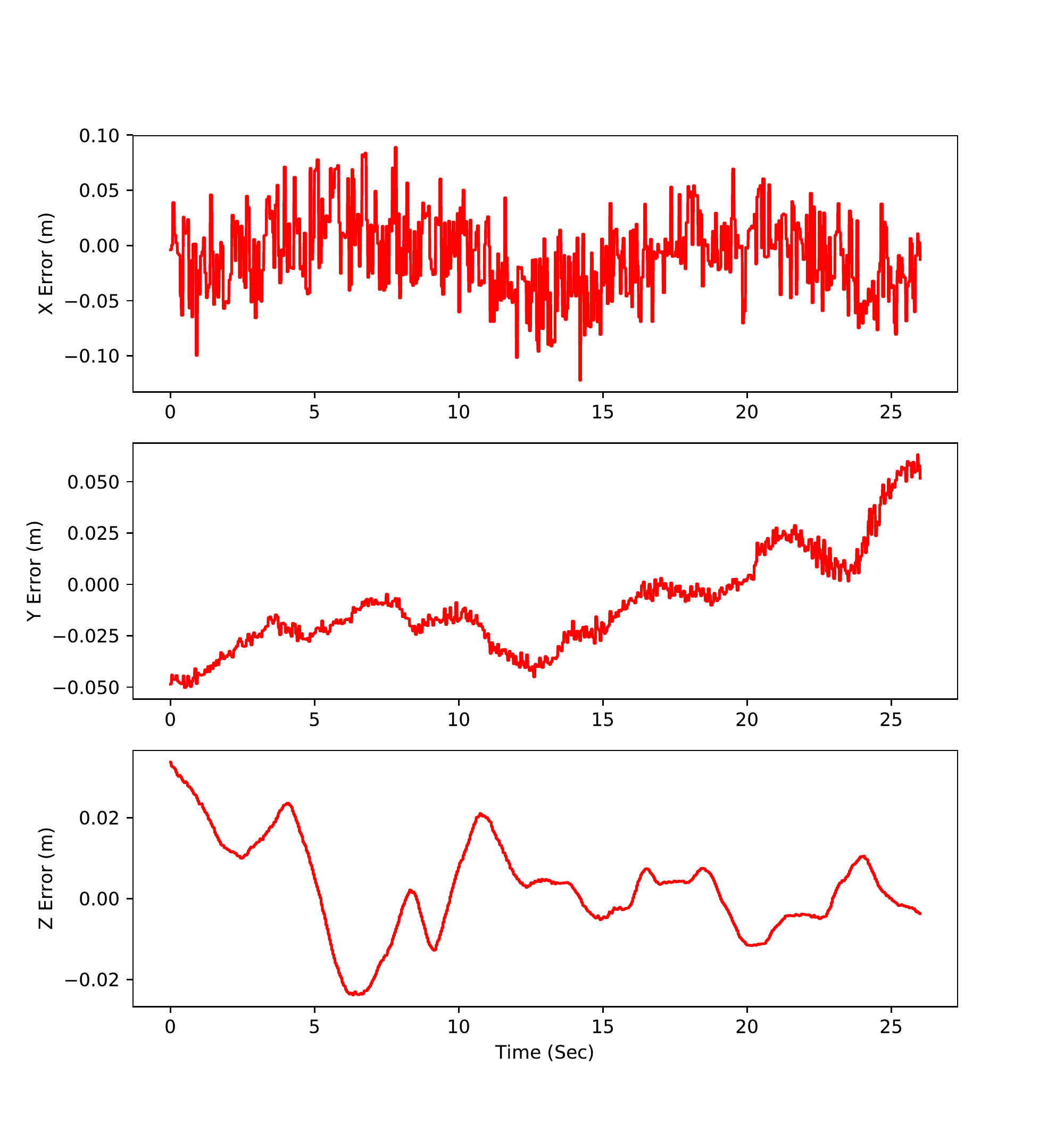}
	\caption{Control errors history during the whole aerial robotic arm moving.}
	\label{fig:rpy}
\end{figure}

\begin{figure}
	\centering
	\includegraphics[width=0.85\linewidth]{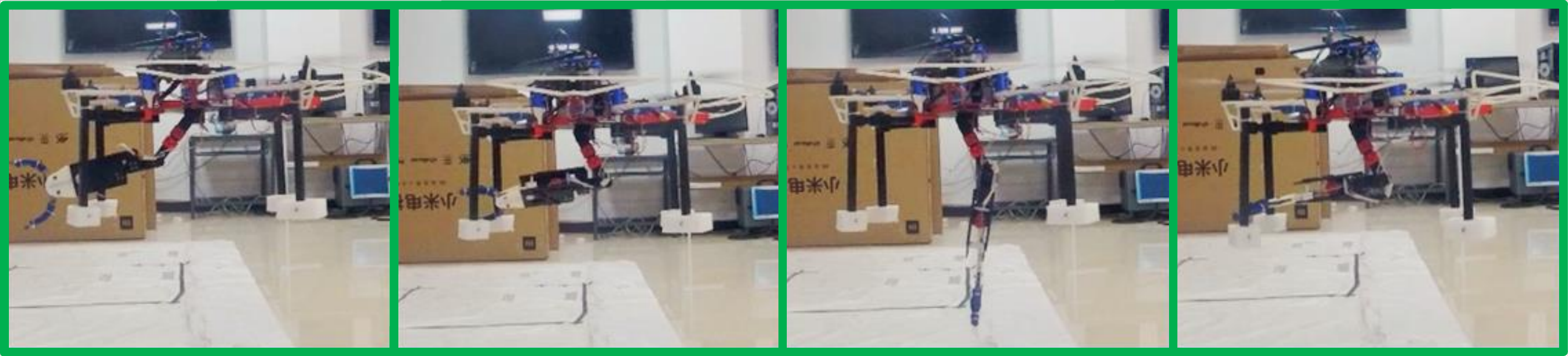}
	\caption{Snapshots of the robotic arm aerial moving experiments. Many movements are carried out during this flight to test the system.}
	\label{fig:aerialmove}
\end{figure}

\subsection{Aerial Robotic Arm Moving Test}
To evaluate the stability and controllability of our system, the UAV was programmed to hovering at a certain point.
In Fig. \ref{fig:rpy}, the control errors of the system during this flight were logged and presented.
After finishing takeoff procedure, the robotic arm will automatically move to multiple points and rotate the end-effector $90^\circ$.
The DCS was activated in this flight, moving the battery to compensate the change of CoG generated by movements of robotic arm.
The standard deviation of control error in world-fixed frame are about 3.64cm in x axis, 2.37cm in y axis and 1.16cm in in z axis, indicated our system can keep hovering at the certain point no matter the robotic arm is moving or not.
we noticed that during the whole test, the error increased at several points including 5, 10 and 22 seconds, respectively. This was caused by the moving of the robotic arm, but our system could always stabilize itself. Fig. \ref{fig:aerialmove} shows some snapshots of the robotic arm aerial moving experiments.

\subsection{Autonomous Grasping}
\begin{figure}
	\centering
	\subfigure[Autonomous grasping of a bottle placed vertically]
	{
		\label{fig:unrotate}
		\includegraphics[width=0.45\linewidth]{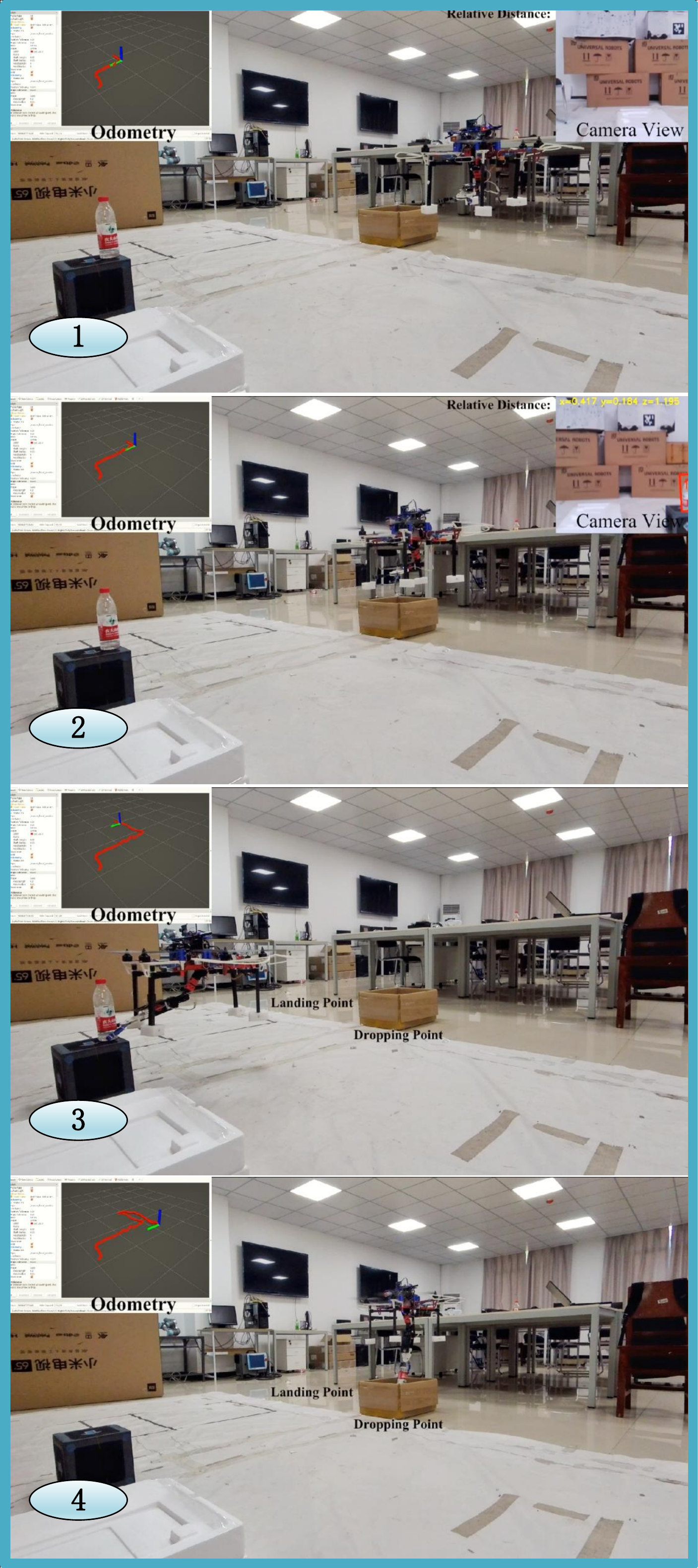}
	}
	\subfigure[Autonomous grasping of a bottle placed obliquely]
	{
		\label{fig:rotate}
		\includegraphics[width=0.45\linewidth]{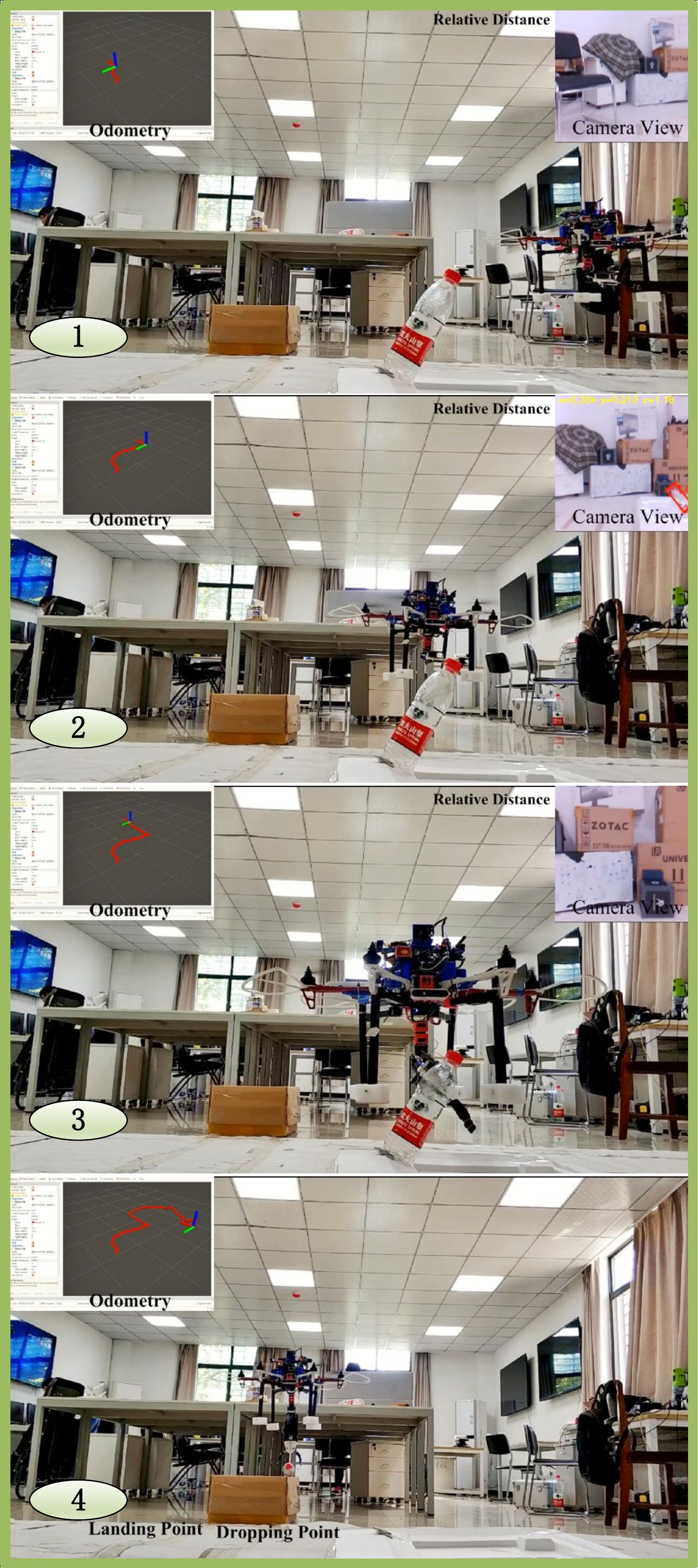}
	}
	\caption{Snapshots of autonomous grasping experiments. The number in subfigures indicate selected moments:
	\ding{172} indicated the UAM is searching specific target;
	\ding{173} indicated the UAM detected the target and gives its relative position;
	\ding{174} indicated the UAM hovering at the grasping position and the end effector is about to grasp the bottle.
	\ding{175} indicated the UAM dropping the grasped target at the dropping points.	
	}
	\label{fig:autograsp}
\end{figure}

Autonomous grasping experiments with objects placed vertically and obliquely have also been conducted.
Fig. \ref{fig:autograsp} presents the key steps of these two grasping experiments.
The subfigures were taken at specific moments ordered by the number in the bottom left corner:
\ding{172} indicated the UAM is searching specific target. Here we use the plastic bottle as target;
\ding{173} indicated the UAM has detected the target and given its relative position. 
The UAM will then align its position to the grasping position;
\ding{174} indicated the UAM hovering at the grasping position and the end effector is about to grasp the bottle.
\ding{175} indicated the UAM dropping the bottle at the dropping point.
The odometry given by the onboard EKF in 100hz rates were simultaneously presented in the top left corner, showing our system can run indoor without using any external visual motion caption system.
The detection results and the calculated relative distance of the target objects were also simultaneously presented in the top right corner during searching.

In Fig. \ref{fig:unrotate}, the target was vertically placed at the top of a cube which weight is about 120g.
By comparing images captured at \ding{172} and \ding{173}, it is easily to know the bottle is not shown in \ding{172} but detected in \ding{173}.
The UAM was using a constant speeded to search in \ding{172}, and speeded up to get close to target after \ding{173}.
The end-effector successfully approached and grabbed the target in \ding{174} verified our methods worked well, and the plastic bottle was not flipped by the downward flow.
In \ding{175} the UAM can automatically drop the target at the dropping point indicates our system is capable of automatically finishing the whole grasping task.

In Fig. \ref{fig:rotate}, the procedures were basically same as mentioned above. 
What different is we placed the empty bottle with about $45^\circ$ rotation by using tapes.
\ding{173} gave the detected target with its relative position and rotated angle.
In \ding{174} the end-effector used the rotation information of the target to successfully grasp the target.

These experimental results have shown that our system has capability to accomplish autonomous grasping mission.

\section{CONCLUSION AND FUTURE WORK}
\label{section:conclusion}

In this letter, we developed an approach to enable a UAM to grasp lightweight objects. 
The key challenges included detecting and locating objects in UAM perspective, CoG compensation, and the flow influence to the objects, all of which made the autonomous grasping mission very difficult. 
We showed the effectiveness of our approach in real tests with the ability to detect, locate and grasp objects with arbitrary pose in GPS-denied environments without relying on the visual motion caption system.
We believe that the proposed solution, both in terms of hardware and algorithms, will be useful in not only the aerial grasping missions but also general grasping missions since the vision system can be applied to any kind of robotic arm. 
Future work will be set out to investigate how to estimate the 6D pose of object in real-time. 
We will also improve the stability by using the servos with torque feedback and adopting more advanced controller for the UAM.

\bibliographystyle{IEEEtran}
\bibliography{root}

\end{document}